**Title: Introducing a Generative Adversarial Network Model for Lagrangian Trajectory Simulation**

**Authors:**

Jingwei Gan[1], Pai Liu[2, *] and Rajan K. Chakrabarty[2, 3]

**Affiliations:**

[1] Department of Chemical and Biological Engineering, Illinois Institute of Technology, Chicago, Illinois 60616, USA

[2] Center for Aerosol Science and Engineering, Department of Energy, Environmental and Chemical Engineering, Washington University in St. Louis, St. Louis, Missouri 63130, USA

[3] McDonnell Center for the Space Sciences, Washington University in St. Louis, St. Louis, Missouri 63130, USA

[*] Address correspondence to: p.liu@wustl.edu

# ABSTRACT

We introduce a generative adversarial network (GAN) model to simulate the 3-dimensional Lagrangian motion of particles trapped in the recirculation zone of a buoyancy-opposed flame. The GAN model comprises a stochastic recurrent neural network, serving as a generator, and a convoluted neural network, serving as a discriminator. Adversarial training was performed to the point where the best-trained discriminator failed to distinguish the ground truth from the trajectory produced by the best-trained generator. The model performance was then benchmarked against a statistical analysis performed on both the simulated trajectories and the ground truth, with regard to the accuracy and generalization criteria.

# I. INTRODUCTION

Lagrangian trajectory simulation is widely applied in multidisciplinary systems to study subjects ranging from Brownian particle diffusion at nanometer scale [1,2] to animal and human movements on miles of landscape [3,4]. The simulated trajectories, from which statistical inferences are often made, provide insights on various topics ranging from particle aggregation kinetics [1,2] to urban planning [3] and search-and-rescue strategies [4,5]. The conventionally adopted first principle method requires well-defined motion equations that govern the movement. When adopting such a method, difficulty commonly arises from the multitude of the motion driving forces [1-5]. Ideally, the model should incorporate all the variables that contribute to the dynamic changes. But a comprehensive inclusion is not feasible when the system is complex, and thus assumptions are made for simplification [1-5]. Another difficulty can arise from the fact that the driving forces of moving objects, especially active objects (e.g. animals and human beings), may not be readily quantifiable and hence very difficult to formally incorporate in the motion equations [3,4]. As an example, a social force model incorporates psychological factors in Langevin equations for modeling human movements [3]. Such a model, however, requires an elaborate design and considerable domain knowledge. Alternatively, statistical approaches have been adopted to address this challenge, such as correlated random walks and Levy flights, in which movements are regulated by probability distribution functions that can be empirically determined [4]. In general, for models based on first principle methods, their performance is limited by many factors, such as domain knowledge, the validity of assumption, model complexity, and the quality of optimization [1-6].

The recent advent of deep learning opened a new avenue for simulating complex systems via a data driven approach [7]. Neural network models have found successful applications in simulating systems that are difficult to tackle with conventional methods [8-10]. For example, a deep generative neural network extracts statistical representation, at multiple levels of abstraction, from experimentally determined datasets (hereafter, the ground truth), and subsequently it generates new instances that share statistical similarity with the ground truth [7, 11, 12]. Among the deep generative neural networks, the variational autoencoder (VAE) [11] and the generative adversarial net (GAN) [12] are the two most widely-applied models. Previously, we explored the possibility of simulating the motion of particles trapped in a buoyancy-opposed flame (introduced in Ref [13-15]) using a VAE model [16]. We found that although the VAE could successfully generate trajectories that were statistically accurate, the model was prone to overfitting the training sets, hence compromising the generalization of the output [16]. Furthermore, the VAE model operates with a deterministic input-output dimension. As a result, the model generates only trajectories with the same length as the model input [16]. In this work, we address these issues encountered in the previous work by introducing a GAN model for trajectory simulation. The GAN has two major components: a generator, which comprises a stochastic recurrent neural network (SRNN) [17, 18], and a discriminator, which is a multilayer convolutional neural network (CNN) [19, 20]. The generator randomly generates trajectory instances, and the discriminator attempts to distinguish the generated trajectories (*fake*) from the ground truth (*real*). Both the generator and discriminator are trained simultaneously, until they are both so developed that the best-trained discriminator fails to discriminate the *real* from the *fake* that are produced from the best-trained generator [7,12].

The rest of paper is organized as follows: In section II, we describe the GAN model architecture, objective function, and training procedure. In section III, we evaluate the GAN model's performance based on accuracy and generalization criteria. We conclude this paper by discussing the advantages of our GAN model, as well as the difficulties that one may encounter when deploying it.

## II. METHODS

### A. Architecture of the GAN model

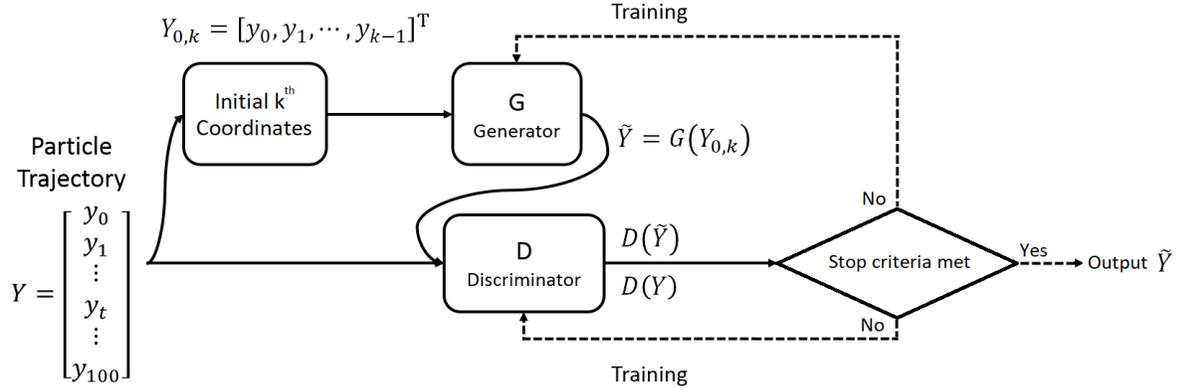

**FIG. 1. Architecture of the generative adversarial network used in this work**

Figure 1 shows the overall architecture of the GAN model, along with the connections among its major components. The model input is 15 experimentally determined particle trajectories, which each of which comprised 1,100 timesteps, corresponding to a total duration $t_n \approx 5.24\,\text{s}$ (Refer to Ref [16] for the experimental acquisition of the particle trajectories). Subsections of Each trajectory containing 100 coordinates was used for training, written as $Y = [y_0, y_1, \cdots, y_t, \cdots y_{100}]^{\text{T}}$ (where $y_t$ represents the particle coordinate at time $t$), and $Y$ serves as the GAN model input per Fig. 1. In the following, we outline the detailed mechanism by which the GAN model functions, following the data flow direction. First, the model samples the first $k^{\text{th}}$ elements from $Y$, concatenating them into an array, $Y_{0,k} = [y_0, y_1, \cdots, y_{k-1}]^{\text{T}}$, which serves as an initial condition for the random trajectory generation. Next, $Y_{0,k}$ is fed to the generator network, which randomly propagates the trajectory datasets into $\tilde{Y} = [y_0, \cdots, y_{k-1}, \tilde{y}_k, \tilde{y}_{k+1}, \cdots \tilde{y}_{100}]^{\text{T}}$.

This generation process is formally written as $\tilde{Y} = G(Y_{0,k})$. Subsequently, the generated trajectory $\tilde{Y}$ and the ground truth $Y$ are sent to the discriminator network, wherein the probability values $D(\tilde{Y})$ and $D(Y)$ are calculated. These two probability values respectively measure the likelihoods that $\tilde{Y}$ and $Y$ are deemed *real* by the discriminator.

In the following sections, we will describe the architecture of the compact SRNN, which serves as the generator, and the CNN, which serves as the discriminator. This description is immediately followed by a paragraph detailing the construction of the objective function and the procedure for the adversarial training.

**B. SRNN serving as generator**

Figure 2 shows a slice of an SRNN network composed of three stochastic recurrent cells (SRC). Such a recurrent connection makes a sequential network structure that has demonstrated special advantages in processing time-series datasets [21-23]. Without losing generality, we describe the function of the SRNN by illustrating how it generates the coordinate at a specific timestep, when given the $k$ preceding coordinates. For example, Fig. 2 shows that the $k$-element array, $[\tilde{y}_{t-2}, \tilde{y}_{t-3}, \cdots, \tilde{y}_{t-k+1}]^{\mathrm{T}}$ is fed to the SRC that calculates $\tilde{y}_{t-1}$. This newly calculated coordinate is concatenated to the preceding ones, and the $k$-element array is updated to $[\tilde{y}_{t-1}, \tilde{y}_{t-2}, \cdots, \tilde{y}_{t-k}]^{\mathrm{T}}$. The updated $k$-element array is then fed to the second SRC in FIG. 2, leading to the generation of $\tilde{y}_t$. This operation repeats itself, generating $\tilde{y}_{t+1}$, and so on. One should note that exceptions occur at the beginning of the trajectory propagation, that is, when $t < k$. At the first timestep ($t = t_0$), the SRC takes $Y_{0,k}$ as its input, generating $\tilde{y}_k$. The $k$-element array is then updated to $[y_1, y_2, \cdots, y_{k-1}, \tilde{y}_k]^{\mathrm{T}}$, which is next used to generate $\tilde{y}_{k+1}$, and so forth. A hundred SRC units were deployed in our generator, matching the length of $Y$ (and $\tilde{Y}$).

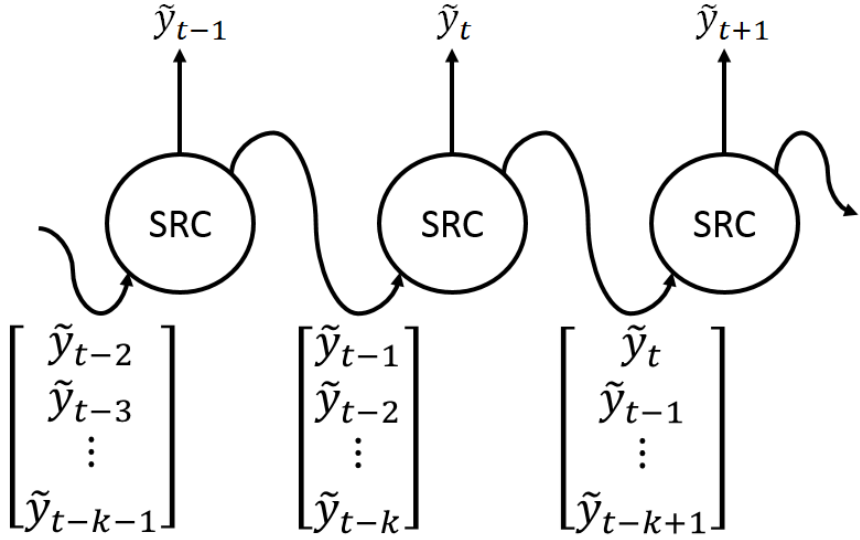

**FIG. 2. Recurrent connection of the SRC units**

The detailed working mechanism of a single SRC unit is shown in FIG. 3. Within the SRC shell, the input $k$-element array, for example, $[\tilde{y}_{t-1}, \ \tilde{y}_{t-2}, \ \cdots, \ \tilde{y}_{t-k}]^{\mathrm{T}}$ is assigned to a tensor notation $\mathrm{X}_0$. This tensor input is sent to multiple layers of the fully connected neural network, whose inputs ($\mathrm{X}_i$) and outputs ($\mathrm{X}_{i+1}$) have the following mathematical relationship:

$$\mathrm{X}_{i+1} = \varphi(w_{i+1} \cdot \mathrm{X}_i + b_{i+1}), \tag{1}$$

where $w_{i+1}$ and $b_{i+1}$, respectively representing the weight and bias matrices of the neuron layer, are the variables to be determined during the model training, and $\varphi()$ is an activation function. The output $\mathrm{X}_1$ of the first layer, after being concatenated by the original input $\mathrm{X}_0$, is then sent to the second layer of the fully-connected neural network, which outputs $\mathrm{X}_2$. Subsequently, the tensor $\mathrm{X}_2$ is simultaneously fed to another two fully-connected neural networks, wherein its mean ($\mu$) and standard deviation ($\sigma$) are individually calculated. The pair of variables $\mu$ and $\sigma$ is next used to randomly generate a future moving increment $\Delta\tilde{y}$ with a unit

Gaussian noise input. Finally, the coordinate $\tilde{y}_t$ is obtained by adding $\Delta \tilde{y}$ to $\tilde{y}_{t-1}$, and the SRC puts $\tilde{y}_t$ at the top of preceding coordinate list, updating the $k$-element array from $[\tilde{y}_{t-1}, \tilde{y}_{t-2}, \cdots, \tilde{y}_{t-k}]^{\mathrm{T}}$ to $[\tilde{y}_t, \tilde{y}_{t-1}, \cdots, \tilde{y}_{t-k+1}]^{\mathrm{T}}$.

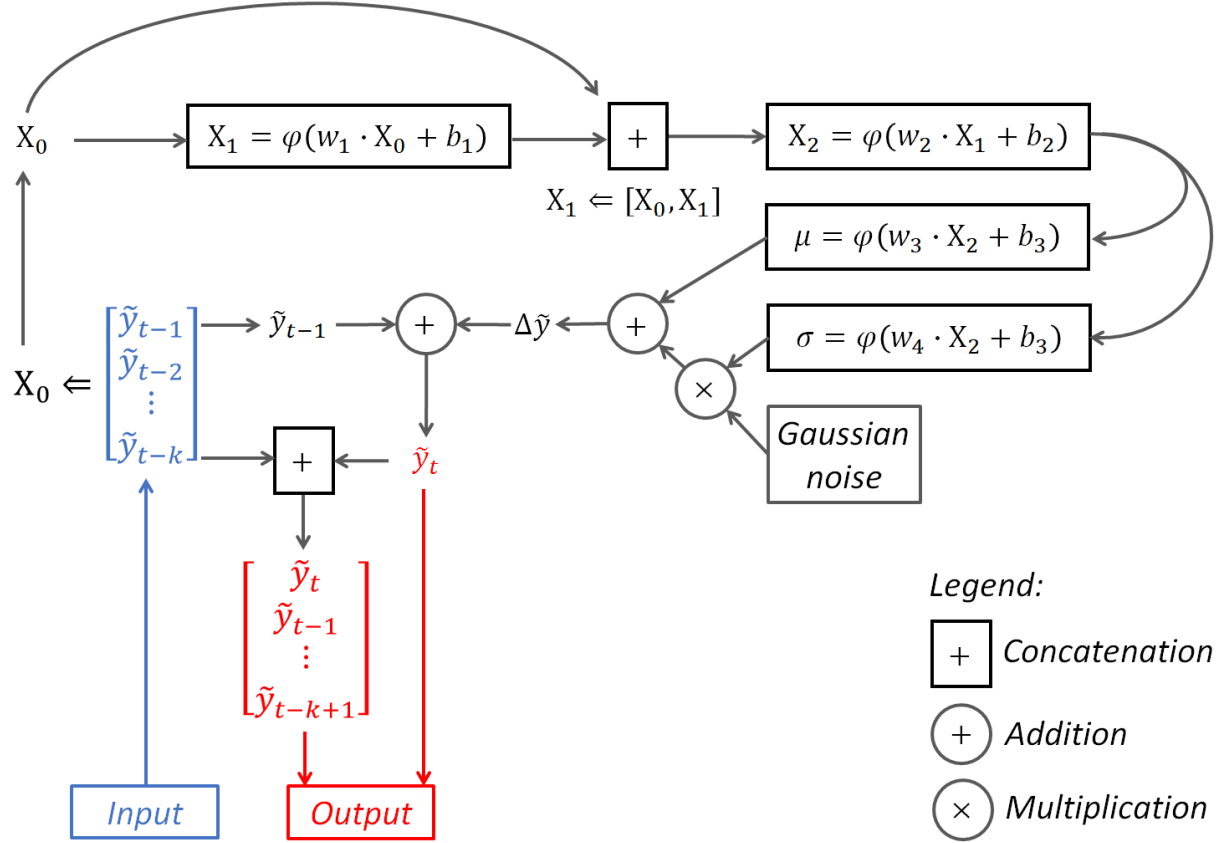

**FIG. 3. Computation graph illustrating the function of a single SRC unit**

Our neural network model differs from the conventional RNN, which follows a deterministic algorithm [21-23]. Randomness is introduced to the model, which accounts for the inherent stochasticity of the particle motion. Also note that our model structure is more compact than the SRNNs introduced in Ref [17, 18], so that it is easier to train and deploy. This concludes our description of the generator network, and in the following section, we move on to the discriminator network.

## C. CNN serving as discriminator

Figure 4 shows how our discriminator network maps the spatially distributed features from the input trajectory datasets ($Y$ and $\tilde{Y}$) to the probability values $D(Y)$ and $D(\tilde{Y})$, by which the ground truth and the generator output are respectively deemed *real*. Three feedforward convoluted neural layers constitute our CNN [19, 20], which recognizes local conjunction features from the input and generates lower dimensional feature maps. Exponential linear units (ELU) are used as the activation functions for these convolutional layers. Dropout with a rate of 0.5 is introduced to avoid overfitting. The output of the third feature map is next passed through a fully connected neural layer with a sigmoid activation function, which calculates the probability $D(Y)$ (or $D(\tilde{Y})$). The CNN, which can detect higher level abstraction, takes fewer variables than the fully-connected neural networks, and thus is quiet well suited as the discriminator [24, 25].

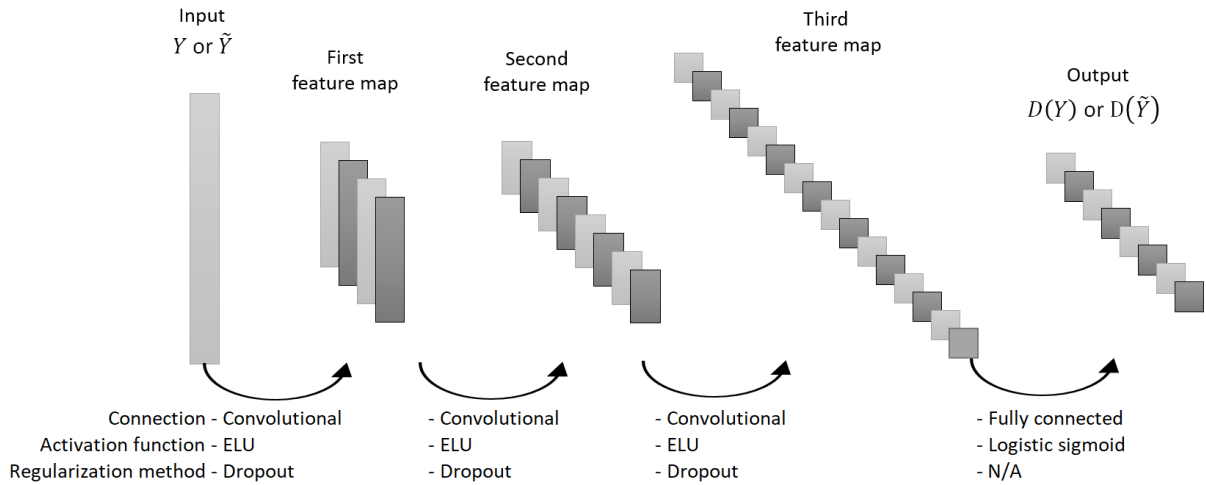

**FIG. 4. Structure of the discriminative CNN**

## D. Objective function and model training

We use a standard cross entropy [12] to measure the cost in the adversarial training of the generator SRNN and the discriminator CNN, and the objective function can be written as

$$J_G = \mathbb{E}\left\{\log\left[1 - D\left(G(Y_{0,k})\right)\right]\right\}, \tag{2.1}$$

$$J_D = -\mathbb{E}\{\log[D(Y)]\} - \mathbb{E}\left\{\log\left[1 - D\left(G(Y_{0,k})\right)\right]\right\}, \tag{2.2}$$

where $\mathbb{E}$ denotes calculating expectation values for the arguments housed in the curly brackets and log is the natural log function. Note that when there is more than one trajectory instance in the training minibatch, the arithmetic mean values of $J_G$ and $J_D$ are calculated and used. The algorithm of the adversarial training is outlined in Table 1:

Table 1. Adversarial training of SRNN model with a minibatch containing $m$ trajectory instances

---

**for** number of training iteration **do**
    **for** $k$ steps **do**
        • Sample a minibatch of $m$ trajectory segments $\{Y^1, Y^2, \ldots, Y^m\}$
        • Take the initial coordinates of each trajectory $\{Y_{0,k}^1, Y_{0,k}^2, \ldots, Y_{0,k}^m\}$
        • Generate $m$ trajectories, $\{\tilde{Y}^1, \tilde{Y}^2, \ldots, \tilde{Y}^m\}$, using the generative SRNN model
        • Update variables in the generator by descending their stochastic gradients $\nabla_{\phi_G} J_G$
        • Update variables in the discriminator by descending their stochastic gradients $\nabla_{\phi_D} J_D$
    **end for**
    **print the** mean values of $J_D$ and $J_G$, and determine if early stop is necessary
**end for**

---

Here, $\phi_G$ and $\phi_D$ are variables involved in the generator and discriminator networks, respectively; $\nabla_{\phi_G} J_G$ and $\nabla_{\phi_D} J_D$ are the gradients of generator loss and discriminator loss with respect to their own parameters, respectively. Model construction and training were conducted using

TensorFlow®. Stochastic gradient decent of the objective function was performed using the Adam optimizer loaded with a constant learning rate. Early stop was used to prevent overfitting and instability [26]. The adversarial training lasted for 5,000 epochs of training, at which point the discriminator returned $D(Y) \approx D\big[G\big(Y_{0,k}\big)\big]$. After the training, the generator iterated every 11 times, producing a trajectory sequence with 1,100 timesteps, matching the length of $t_n$.

## III. RESULTS AND DISCUSSION

Figure 5 compares the trajectories generated using the GAN model to the trajectories determined in experiments. Qualitatively speaking, good similarity can be observed between the GAN output and the ground truth. Also note that 3-dimensional cylindrical coordinates best describe our system, that is $y = [r, \theta, z]$.

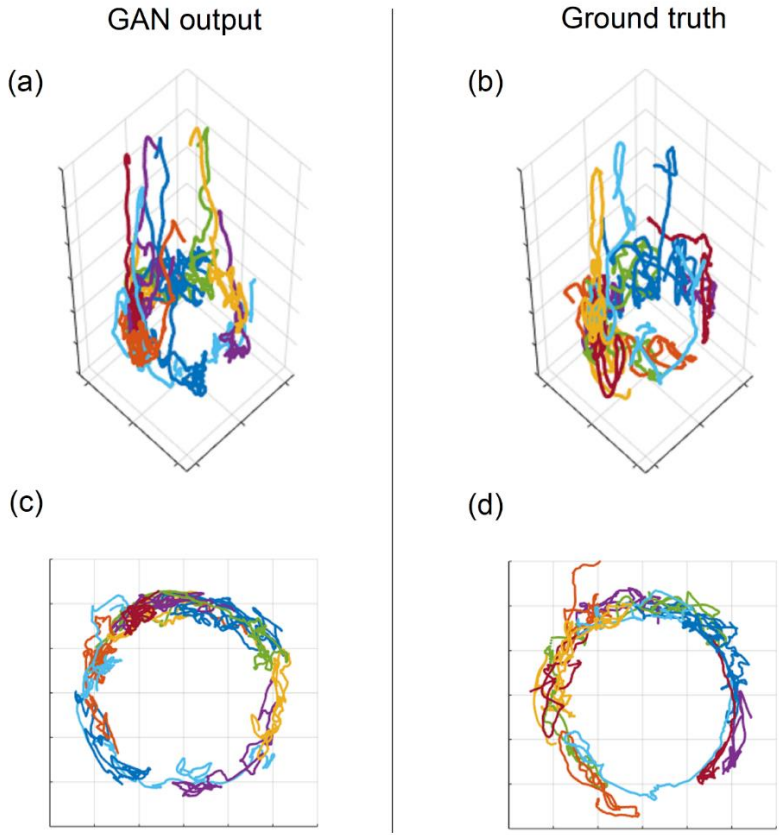

**FIG. 5. Comparison between the GAN model output and the ground truth. (a) and (b) respectively show the 3-dimensional trajectories generated using the GAN model and that determined in experiments. The distance between tick marks corresponds to a length of 50 mm. (c) and (d) show the trajectories projected to the horizontal plane. The distance between tick marks corresponds to a length of 20 mm. In each panel, 15 individual trajectories with 1,100 timesteps are plotted.**

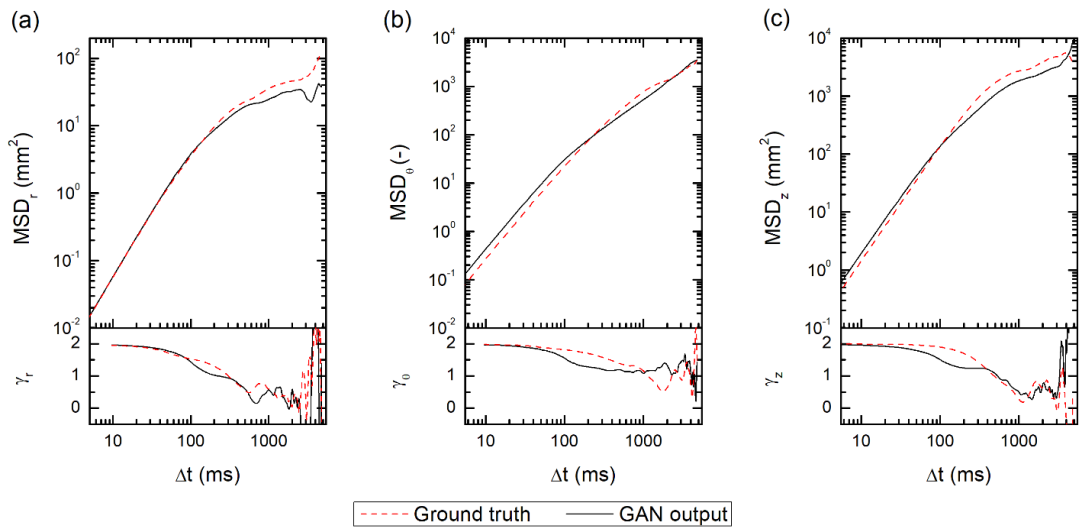

**FIG. 6. The power-law scaling relationship $MSD \propto \Delta t^{\gamma}$ for the motions in the (a) $r$, (b) $\theta$, and (c) $z$ directions. The time-evolution behaviors of the scaling exponent $\gamma$ are shown in the subpanels**

We next discuss the performance of GAN model with regard to the accuracy of its output. Figure 6 shows the scaling analysis performed on the mean squared displacements (MSD) of the GAN outputs and the ground truth. Here, the $\text{MSD}^i$ of each trajectory was calculated with a changing timescale ($\Delta t$) in a time-averaged manner [27, 28]:

$$MSD^i(\Delta t) = \frac{t_0}{t_n - \Delta t} \sum_{t=t_0}^{t_n - \Delta t} \left(y_{t+\Delta t}^i - y_t^i\right)^2, \tag{3.1}$$

$$\text{and } MSD = \frac{1}{15} \sum_{i=1}^{15} MSD^i. \tag{3.2}$$

Note that $t_0$ also represents a unit time interval $\approx 4.76$ms, and the $\Delta t$ in Eq. (3.1) only takes discrete values divisible by $t_0$. Without losing generality, Eq. (3.1) uses $y$ to denote the cylindrical coordinates [$r, \theta, z$], but in reality, we calculate [$MSD_r, MSD_{\theta}, MSD_z$] separately. We observe excellent agreements between the GAN output and ground truth in Fig. 6. The GAN output also accurately mimics the long-term behaviors observed for the ground truth. For example, the time-evolution of the scaling exponent $\gamma$ that parameterizes the power-law relationship:

$$MSD \propto \Delta t^{\gamma}. \tag{4}$$

When $\gamma$ takes values of 2, 1, and 0, the motion respectively manifests as ballistic, diffusive, and trapped [27]. Note that the motion in a direction starts out ballistically $\gamma = 2$. Over time the values of $\gamma$ in the $r$ and $z$ directions approach zero, because the particles are trapped in flame vortices with finite length scale [13]. The values of $\gamma$ in the $\theta$ direction, however, approaches unity at large timescales, indicating normal diffusion, because the angular rotation is unbounded and the correlations in the motion are eventually washed away. Figure 6 compares the time evolutions of $\gamma$ for the simulations and the ground truth, good agreement is observed in all directions. Figure 7 shows the probability distributions of velocities $p(v_r)$, $p(v_\theta)$, and $p(v_z)$ for the GAN outputs and the ground truth. Deviations are negligible.

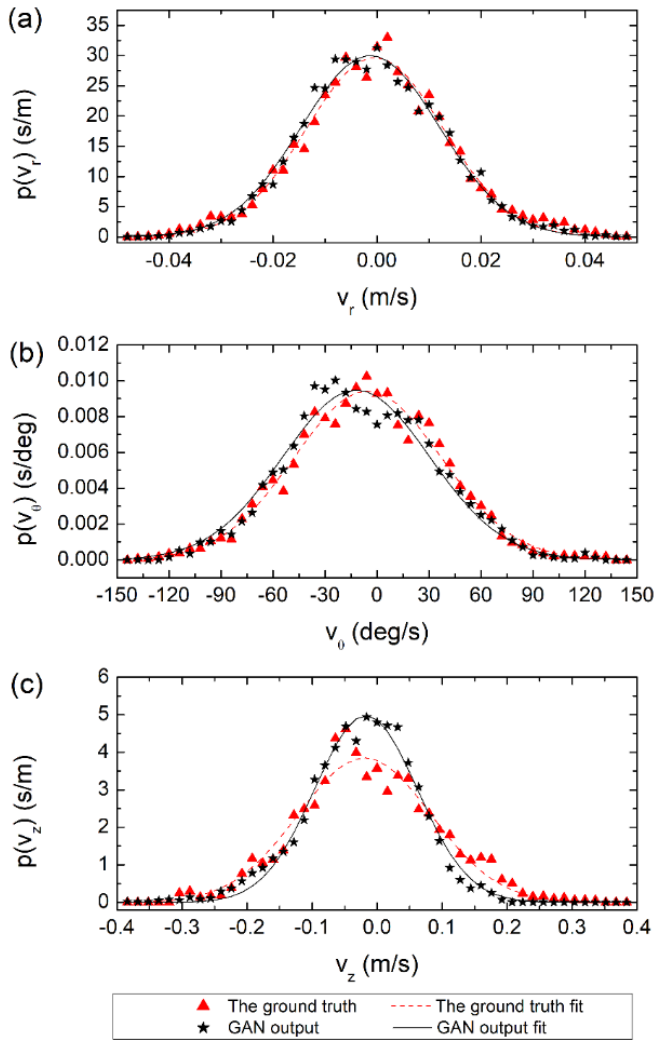

**FIG. 7. Velocity distribution in the (a) $r$, (b) $\theta$, and (c) $z$ directions. Least square fittings following a normal distribution function are performed on the datasets of both the GAN output and the ground truth, shown by the lines**

We next discuss the generalization of the GAN outputs. The degree of generalization is quantitatively assessed using the Pearson correlation coefficients between, for example, the ground truth $Y^i$ and the GAN model output $\tilde{Y}^i$:

$$\zeta\left(Y^i, \tilde{Y}^j\right) = \frac{\mathbb{E}\left[\left(Y^i - \mathbb{E}[Y^i]\right)\left(\tilde{Y}^j - \mathbb{E}[\tilde{Y}^j]\right)\right]}{\mathbb{E}\left[\left(Y^i - \mathbb{E}[Y^i]\right)^2\right]^{1/2} \mathbb{E}\left[\left(\tilde{Y}^j - \mathbb{E}[\tilde{Y}^j]\right)^2\right]^{1/2}}. \tag{4}$$

The condition of $\zeta\left(Y^i, \tilde{Y}^j\right) \approx 1$ indicates $\tilde{Y}^j$ to be highly linearly-correlated with $Y^i$, and zero indicates that they are independent. Note that the calculation of $\zeta\left(Y^i, \tilde{Y}^j\right)$ was individually done using the three cylindrical components of $Y^i$ (and $\tilde{Y}^j$), and the arithmetic mean values of the $\zeta$ in $r$, $\theta$, and $z$ spaces were taken to represent correlation in 3-dimensional space, which are plotted in FIG. 8. Panel (a) shows the self-correlation matrix calculated using the experimental datasets, that is, $\zeta\left(Y^i, Y^j\right)$. Such a self-correlation matrix could be regarded as a baseline for discussing generalization. For example, the elements on the main diagonal ($i = j$), indicating correlation values between identical events, always take a value of unity (white color). The elements *not* on the main diagonal ($i \neq j$), indicating correlation values between independent events, range between *ca.* -0.5 and 0.5, while the population clusters at zero. FIG. 8 (b) shows the correlation $\zeta\left(Y^i, \tilde{Y}^j\right)$ between the ground truth and GAN outputs, and (c) shows the correlation between the ground truth and previous VAE outputs of comparable accuracy [16]. Significant improvement in the degree of generalization can be seen in the GAN model outputs. As the white colored *hotspots* (where $\zeta$ is near unity) prevailing in the correlation matrix of the VAE are not seen in that of the GAN. This observation indicates that the GAN model succeeds in generating statistically accurate trajectories which, at the same time, manifest a sufficient degree of diversity, so that the outputs can be regarded as new events independent of model inputs.

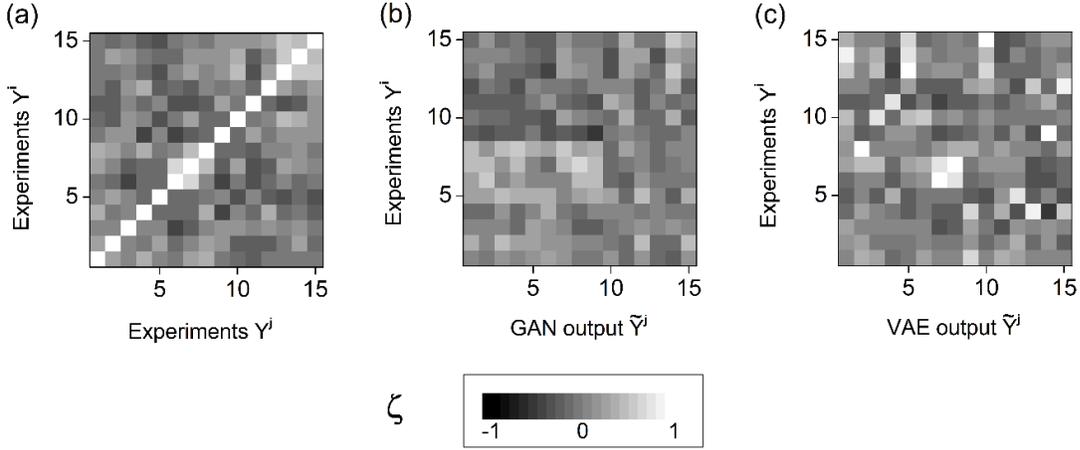

$\zeta$

**FIG. 8. Contour plots for the matrices of Pearson correlation coefficients. (a) shows the self-correlation $\zeta(Y^i, Y^j)$ calculated using 15 experimentally obtained trajectory datasets. (b) shows the correlation $\zeta(Y^i, \widetilde{Y}^j)$ between 15 experimental datasets and 15 trajectory datasets generated using the GAN model introduced in this work. (c) shows the correlation $\zeta(Y^i, \widetilde{Y}^j)$ between 15 experimental datasets and 15 trajectory datasets generated using the VAE model introduced in Ref [16].**

Finally, we comprehensively evaluate the performance of our deep learning models. Following the procedure outlined in Ref [16], we score the accuracy of model outputs according to the statistics related to the spatiotemporal scaling relationship:

$$Accuracy = 1 - \langle \eta \rangle \tag{5.1}$$

with, $\langle \eta \rangle = \frac{t_0}{t_n - t_0} \sum_{\Delta t = t_0}^{t_n} \left| \frac{\log_{10} MSD_Y(\Delta t) - \log_{10} MSD_{\widetilde{Y}}(\Delta t)}{\log_{10} MSD_Y(\Delta t)} \right|. \tag{5.2}$

Note that in (5.1) the values of MSD in three dimensional spaces are used. On the other hand, we score the generalization of the model outputs using the maxima that appear in the Pearson correlation matrices:

$$Generalization = 1 - \langle \zeta \rangle \tag{6.1}$$

with, $\langle \zeta \rangle = \frac{1}{m} \sum_{j=1}^{m} \left[ \max_i \zeta(Y^i, \widetilde{Y}^j) \right]. \tag{6.2}$

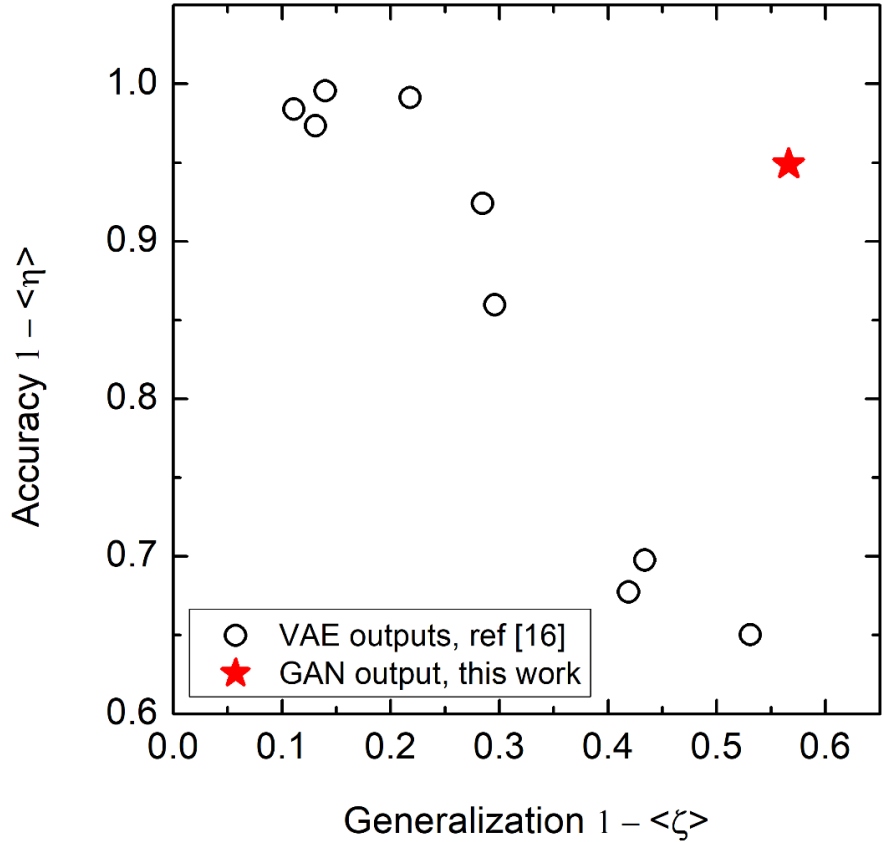

**FIG. 9. A comprehensive evaluation of model performance. The GAN model (star) is scored for accuracy and generalization. For comparison, the performance of the VAE model introduced in our previous work [16] is also plotted (circles).**

Figure 9 compares the performance of the GAN model to that of the VAE model introduced in Ref [16], using accuracy and generalization as criteria. Unlike our VAE model, which operates by minimizing the Euclidean norm between $Y$ and $\tilde{Y}$ [16], the adversarial training scheme used in the GAN is faithful to the statistical representation of the ground truth, without overfitting input data. The superior performance of GAN model also stems from the application of SRNN as the generator, which is a suitable choice for propagating trajectory datasets of sequential nature. We also emphasize that the length of the GAN output is not limited by the length of the input. After being successfully trained, the generator can iterate for an arbitrary number of times, thus propagating the output trajectory to unlimited length (within a scope that makes physical sense).

The difficulty involved in the application of our GAN model resides in the complexity of neural network architecture. Unlike feedforward networks, the SRNN generator contains hundreds of SRC units. Training such an intricate network demands considerable computation power, and one also needs to tackle the resulting gradient vanishing or exploding problems [29]. The delicate trade-off between model complexity and model performance should be taken into consideration. For example, reducing the number of SRC units in the SRNN alleviates the computational load. However, this treatment requires the model inputs to be sliced into correspondingly shorter segments, and the GAN model may fail to perceive the long-term statistical representations in the ground truth. In this work, we deploy 100 SRC units in the generator, striking a balance between load and performance. This hyper-parameter should be carefully chosen, according to the system being studied, the application, and the computational resources. Other difficulties are the ones commonly encountered in adversarial training, such as training instability and mode collapsing, which require fine-tuning by trial and error until a stable model configuration is reached [26, 30-32].

## IV. CONCLUSION

In summary, our GAN model has achieved state-of-the-art performance in simulating the Lagrangian trajectories of particles trapped in the recirculating zone of a buoyancy-opposed flame. Scaling analysis shows that the 3-dimensional MSD of the simulated motion deviates by only ca. 5% from that of the ground truth. The adversarial training scheme manages to transfer the statistical representation from the input without overfitting the data, thus the generalization of the model output is guaranteed. Using an SRNN as the generator, the GAN model can freely propagate the simulated trajectory without such limitations as the model input-output dimension.

**ACKNOWLEDGEMENTS**

This research is funded by US National Science Foundation (NSF) grants AGS-1455215 and CBET-1511964, and the NASA Radiation Science Program grant NNX15AI66G.